\documentclass[11pt]{article}

\usepackage[final]{acl}

\usepackage{times}
\usepackage{latexsym}

\usepackage[T1]{fontenc}

\usepackage[utf8]{inputenc}

\usepackage{microtype}

\usepackage{inconsolata}

\usepackage{graphicx}

\usepackage[capitalise]{cleveref}
\usepackage{subcaption}
\usepackage{booktabs}

\everypar\expandafter{\the\everypar\looseness=-1}
\title{Titans Revisited: A Lightweight Reimplementation and Critical Analysis of a Test-Time Memory Model}

\author{Gavriel Di Nepi \\
Sapienza University of Rome \\
\small\texttt{dinepi.2067753@studenti.uniroma1.it} \\\And
Federico Siciliano \\
Sapienza University of Rome \\
\small\texttt{siciliano@diag.uniroma1.it} \\\And
Fabrizio Silvestri \\
Sapienza University of Rome\\
\small \texttt{fsilvestri@diag.uniroma1.it} \\
}

\begin{document}
\maketitle
\begin{abstract}
By the end of 2024, Google researchers introduced \textit{Titans: Learning at Test Time}, a neural memory model achieving strong empirical results across multiple tasks.
However, the lack of publicly available code and ambiguities in the original description hinder reproducibility.
In this work, we present a lightweight reimplementation of Titans and conduct a comprehensive evaluation on Masked Language Modeling, Time Series Forecasting, and Recommendation tasks.
Our results reveal that Titans does not always outperform established baselines due to chunking. However, its Neural Memory component consistently improves performance compared to attention-only models. These findings confirm the model’s innovative potential while highlighting its practical limitations and raising questions for future research. 
\end{abstract}

\section{Introduction}
Natural Language Processing (NLP) has advanced rapidly alongside the rise of deep learning. One major breakthrough was the introduction of the Transformer architecture by \citet{vaswani2017attention}, which enables parallel processing of sequences, leading to substantial gains in training efficiency and model performance. Transformers are the foundation of Large Language Models (LLMs), such as GPT \cite{brown2020language} and LLaMA \cite{touvron2023llama}. These models are trained on massive datasets and demonstrate impressive zero-shot and few-shot learning capabilities.
However, these models rely on a fixed-length context window for in-context learning, which creates challenges when handling long sequences due to quadratic computational cost and limited memory capacity.

Recent efforts to overcome these limitations fall into three categories: improved attention mechanisms, model adaptation at inference time, and alternative attention mechanisms.
In the first category, Longformer \cite{beltagy2020longformer} introduces sparse attention to enable more efficient handling of longer sequences without sacrificing accuracy. More recently, \citet{fountas2024human} propose a system inspired by human episodic memory that leverages retrieval, clustering, and scoring to enable effectively infinite context modeling. 
A second line of work focuses on Test-Time Learning (TTL), which enables models to adapt during inference via gradient-based updates.
\citet{sunlearning} propose recurrent models with neural hidden states that are updated using a self-supervised reconstruction loss dynamically as inputs are processed \cite{sun2023learning}.
The third approach abandons attention entirely. Mamba \cite{gu2023mamba} uses state space models (SSMs) to process sequences in linear time through implicit recurrence and gated updates. This allows for efficient long-range modeling without attention.
Titans \cite{behrouz2024titans} combines elements from the first two approaches. It incorporates persistent tokens and a memory module that updates during inference, alongside chunk-based input. However, some architectural details are ambiguous, and the lack of an official implementation complicates reproducibility and independent evaluation.

In this work, we address these ambiguities by providing a lightweight reimplementation of Titans that makes explicit the design choices left unspecified in the original article. We also extend its evaluation to three tasks: (i) Masked Language Modeling, which builds on the original NLP experiments; (ii) Time Series Forecasting, which uses the same dataset for direct comparison; and (iii) Recommendation, a novel domain not included in the original study that allows us to explore the generalization of Titans' memory mechanisms.


Our results show that Titans does not always outperform the baselines, mainly due to inputs chunking. However, its Neural Memory consistently enriches information beyond attention-only models, leading to performance improvements. These findings confirm the model's innovative potential while highlighting unresolved design questions and avenues for further research.

\section{Methodology}
This work presents a reimplementation and analysis of Titans, a model designed for learning at test time that combines attention with neural memory. Although the original publication introduced a promising approach for handling extremely long sequences, the lack of accompanying code and several under-specified design choices made it difficult to reproduce the results. Our methodology has two main objectives: (1) to provide a clear, lightweight implementation of Titans that can serve as a reliable reference, and (2) to systematically evaluate the architectural ambiguities left unresolved in the original description.

\subsection{Titans Architecture}
Titans integrates three complementary components:
\begin{enumerate}
    \item The Core module: a Transformer with local attention that functions as short-term memory. By restricting attention to a local window, this module reduces the quadratic cost of standard self-attention while capturing dependencies within each chunk.
    \item The Neural Long-Term Memory module: the central innovation of Titans. It is a differentiable memory that is updated continuously at inference time via a surprise-driven rule. Decay factors, momentum, and gating coefficients determine when and how information is stored or forgotten, ensuring stability while adapting to new inputs. 
    \item Persistent Memory tokens: task-specific embeddings prepended to each sequence. These tokens remain fixed during inference and act as anchors for attention, injecting prior knowledge and stabilizing the model across multiple chunks.
\end{enumerate}

The original paper proposes three memory integration strategies: as additional context (MAC), as a gating mechanism (MAG), or after the logits (MAL). Our implementation primarily focuses on the MAC variant, in which memory retrieval and persistent tokens are concatenated with the current chunk before attention. We also explore a simplified variant with only the Long-term Memory Module (LMM), which remove attention entirely in order to isolate the contribution of memory.

\subsection{Ambiguities in the Original Paper}
Several crucial details in the original work were not fully specified, making reproducibility challenging. For example, it was unclear whether predictions should be generated from only the last chunk or all chunks, which affects memory requirements and context retention.
Another issue was the dimensionality reduction strategy after concatenating memory and persistent tokens, i.e. whether the outputs should include all tokens or be projected back to the original chunk size.
There was further ambiguity about whether the MAC block is a shallow encoder or a stackable layer, which affects how the model scales with depth.
Details of the internal attention structure, such as the number of heads, layers, and positional encoding, were also omitted, limiting a fair comparison with the baselines.
Finally, the specific contribution of test-time updates was not isolated, making it difficult to assess whether improvements stemmed from dynamic memory adaptation or other factors.

\begin{table*}[!ht]
\centering
\begin{tabular}{l|llllll}
\toprule
\textbf{} & HR@1 & \textit{HR@5} & HR@10 & NDCG@5 & NDCG@10 & MRR \\
\midrule
Bert4Rec & \textbf{0,0606} & \textbf{0,1758} & \textbf{0,2603} & \textbf{0,2123} & \textbf{0,1984} & \textbf{0,4451} \\
\midrule
MAC & \underline{0,0464} & \underline{0,1434} & \underline{0,2177} & \underline{0,2018} & \underline{0,1689} & \underline{0,4371} \\
MAC only attention & 0,0412 & 0,1404 & 0,2137 & 0,1596 & 0,1456 & 0,3418 \\
\bottomrule
\end{tabular}
\caption{Results on the MovieLens 1M Recommendation System task, comparing BERT4Rec with Titans in its MaC variants. The best values for each metric are highlighted in bold, while the second-best results are underlined.}
\label{tab:recsys}
\end{table*}

\subsection{Our Approach}
To address these issues, we design a modular reimplementation of Titans in which all mechanisms are explicitly defined and configurable. For each ambiguity, we implement and empirically compare multiple plausible strategies. For instance, we test alternative output fusion methods between attention and memory, as well as different dimensionality reduction techniques. Memory retrieval, update and integration are fully transparent to ensure reproducibility and extensibility of our results.
Our implementation emphasizes interpretability and empirical rigor. The architecture is intentionally minimal, remaining faithful to the original description without introducing undocumented heuristics. We conduct controlled comparisons with baseline Transformers of similar computational cost, as well as ablation studies to isolate the contributions of chunking, persistent tokens, and neural memory. 

In summary, our methodology balances fidelity to the original proposal with transparency and empirical rigor. By formalizing Titans’ components, explicitly resolving ambiguities and providing a reproducible implementation, this work establishes a robust framework for analyzing its potential in long-context reasoning and test-time learning.

\subsection{Experimental setup}
We evaluate our reimplementation of Titans across three tasks: Masked Language Modeling (MLM), Time Series Forecasting and Recommendation. Together, these tasks capture symbolic reasoning over text, and long-range temporal dependencies and shorter user–item interaction sequences. This allows us to assess the behavior of the model in diverse contexts.

For MLM, we use the CC-News dataset \cite{Hamborg2017}, which consists of over 700,000 English news articles, and compare Titans with a BERT-like \cite{devlin2019bert} baseline that was trained with the same computational budget. To evaluate the contribution of individual components, we conduct ablation studies by selectively removing persistent tokens and neural memory.

For time series forecasting, we adopt the widely used ETTh1 dataset of transformer temperature readings \cite{zhou2021informer}, and compare Titans with two strong baselines: a Long Short-Term Memory (LSTM) model and iTransformer \cite{liu2023itransformer}. This task primarily tests the standalone ability of the neural memory.

For the recommendation task, we use the MovieLens 1M \cite{harper2015movielens} dataset with a next-item prediction setup, in which the model predicts the next movie based on the user history. As with MLM, we perform an ablation study, and compare Titans against a BERT4Rec \cite{sun2019bert4rec} baseline.

All models are trained with comparable parameter counts and optimization settings. Due to hardware constraints, input sequences are divided into chunks (typically 32–128 tokens), which enables us to test the memory's ability to bridge discontinuous segments. Across tasks, we systematically vary context length (128-512) and memory usage to evaluate robustness.

\section{Results}

\begin{table}[!ht]
\resizebox{\columnwidth}{!}{
\begin{tabular}{l|ll}
\toprule
\textbf{} & Accuracy& F1 Score\\
\midrule
Baseline& \underline{0,0558} & 0,0420\\
\midrule
MAC & 0,0551& \textbf{0,0431}
\\
MAC only attention & 0,0539& 0,0400\\
MAC without persistent & \textbf{0,0559}& \underline{0,0429}
\\
MAC without memory & 0,0523& 0,0399
\\
\bottomrule
\end{tabular}
}
\caption{Ablation study on the MLM task with \textit{sequence length} = 128 and \textit{chunk size} = 32. The comparison highlights the contribution of different Titans components. Best results for each metric are highlighted in bold, while the second-best are underlined.} 
\label{tab:ablation}
\end{table}

\begin{figure*}[t]
  \centering
  \begin{subfigure}{0.48\linewidth}
    \includegraphics[width=\linewidth]{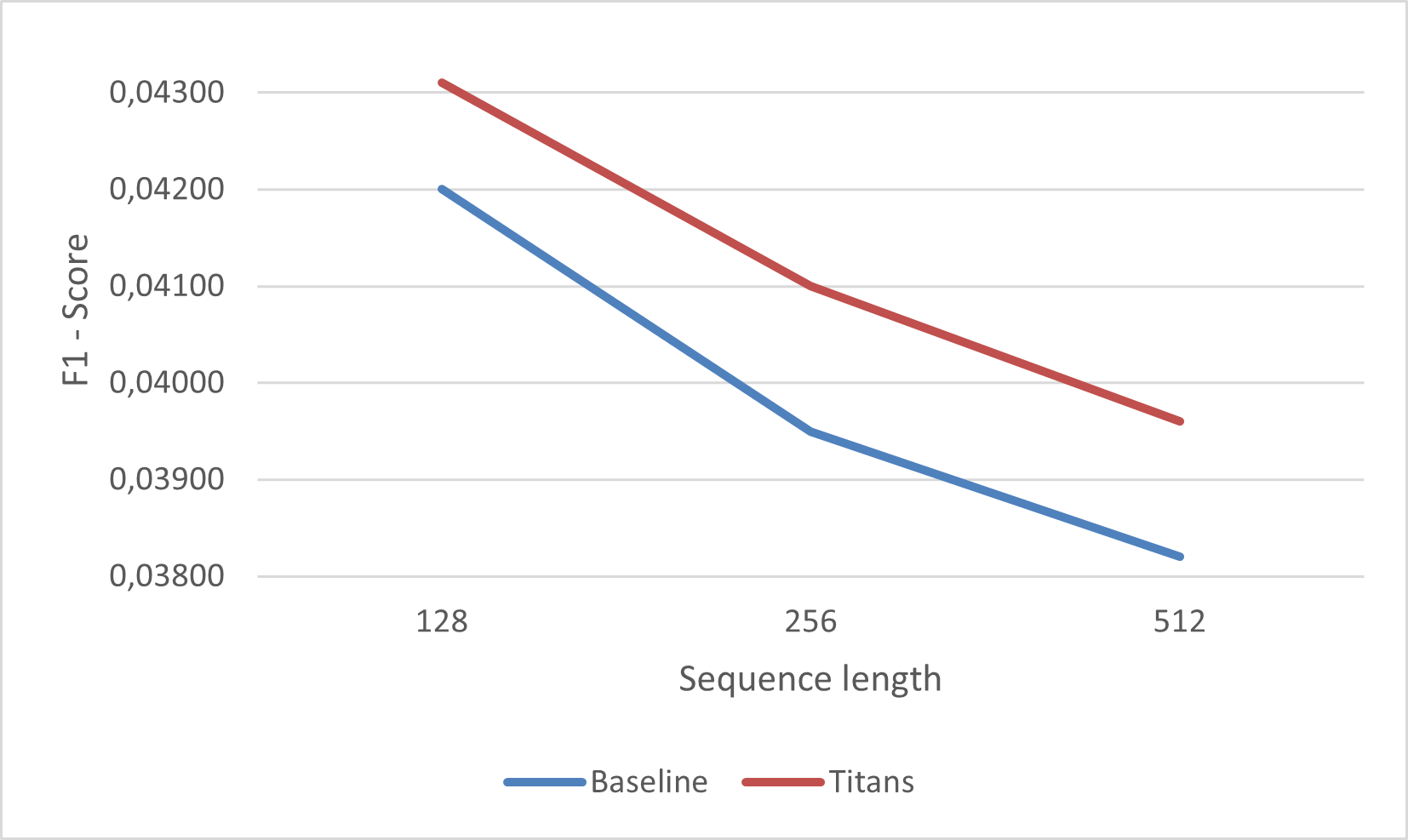}
    \caption{Effect of sequence length on F1 - Score}
    \label{fig:seq-length}
  \end{subfigure}\hfill
  \begin{subfigure}{0.48\linewidth}
    \includegraphics[width=\linewidth]{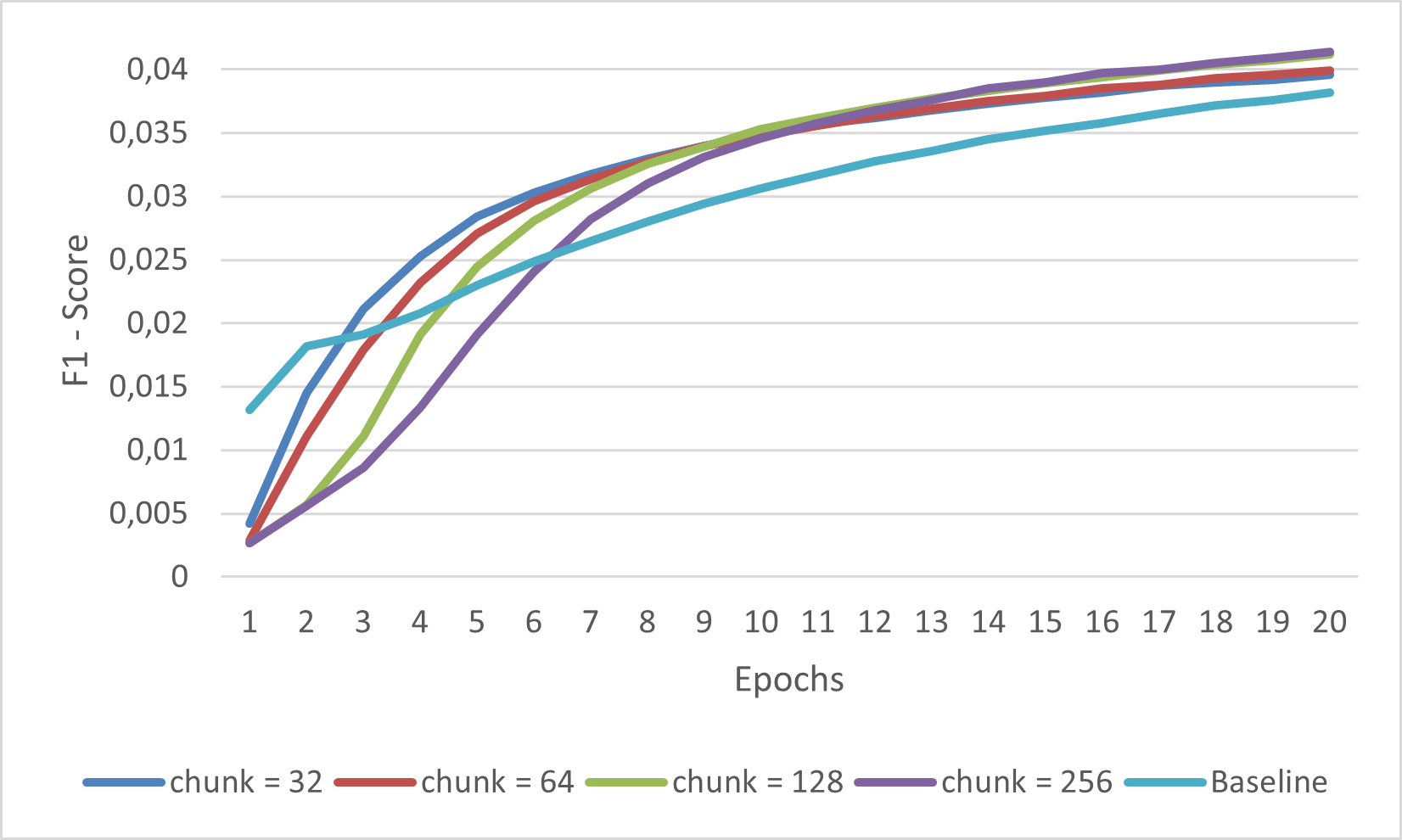}
    \caption{Effect of chunk size on F1 - Score}
    \label{fig:chunk-size}
  \end{subfigure}
  \caption{Analysis of Titans under different input configurations. The plots illustrate how architectural parameters impact model scalability and memory reliance}
  \label{fig:chunk-vs-length}
\end{figure*}

\paragraph{Masked Language Modeling} In the MLM task, Titans matches or surpasses the BERT-like baseline with the same training budget. As shown in \cref{tab:ablation}, neural memory is a decisive factor in achieving these improvements. In contrast, persistent tokens alone have negligible or even negative effects.
This suggests that it is adaptive memory updates rather than static embeddings that dive improvements in Titans.

\paragraph{Recommendation} On MovieLens 1M, Titans failed to outperform the BERT4Rec baseline. However, \cref{tab:recsys} demonstrates the benefit of memory. Specifically, Mean Reciprocal Rank (MRR) increased by $0.09$ with memory (from $0.34$ to $0.43$), when memory was included. Despite the remaining gap with BERT4Rec, this shows that memory mitigates the negative effects of chunking in sequential recommendation and improves modeling user–item dependencies across longer interaction histories.

\paragraph{Time series forecasting} For the time series forecasting task, the memory-only variant LMM matches or exceeds the performance of the iTransformer and LSTM baselines (see \cref{tab:time-series-forecasting}). This suggests that neural memory can independently capture long-term temporal dependencies. However, the metrics do not reach the levels reported in the original Titans paper, indicating that further optimization is required. This gap highlights the limitations of our lightweight implementation and Titans' sensitivity to hyperparameter tuning.

\begin{table}[!ht]
\centering
\begin{tabular}{l|ll}
\toprule
\textbf{} & MSE& MAE\\
\midrule
iTransformer& \underline{0.4925} & \underline{0,5137}\\
LSTM & 0,5524 & 0,5219\\
\midrule

LMM & \textbf{0,4872} & \textbf{0,4837} \\
\bottomrule
\end{tabular}
\caption{Results on the ETTh1 Time Series Forecasting task, comparing iTransformer, LSTM and LMM. The table reports MSE and MAE scores, where the best values are highlighted in bold and the second-best are underlined} 
\label{tab:time-series-forecasting}
\end{table}

\paragraph{Analysis of Context Length and Chunking}
In addition to evaluating Titans' performance on specific tasks, we investigate how the model behaves under different sequence lengths and chunk sizes. These factors are crucial for scalability, as longer sequences exacerbate the limitations of attention, while smaller chunks increase reliance on memory to capture cross-segment dependencies.
As \cref{fig:seq-length} shows, both Titans and the baseline exhibit a decrease in F1 score as sequence length increases, likely due to the increased number of masked tokens.
For the baseline, the attention mechanism must handle more tokens at once.
For Titans, longer sequences create more chunks, placing a greater strain on the memory module, which must store and retrieve a larger amount of information.
Nevertheless, Titans consistently outperforms the baseline at all tested lengths.
All of these results are based on experiments with a chunk size of 32. For example, a sequence of length 128 is divided into four chunks; a sequence of length 512 is divided into 16 chunks.

Since chunking is identified in \cref{tab:ablation} as the main source of performance degradation, we examine the impact of different chunk sizes in \cref{fig:chunk-size}. The results confirm that larger the chunk sizes lead to better performance, albeit at a higher computational cost. This indicates that Titans' potential is likely to be greater than what observed in \cref{fig:chunk-size}.
It is also important to note how the MAC variant of Titans computes attention. The attention module processes not only the current chunk, but also the concatenated retrieved memory and the persistent tokens. Consequently, the effective sequence length is twice the chunk size plus the persistent tokens.
For example, with a chunk size of 256, the computational cost of calculating each attention matrix would be the same as the baseline, but it would need to be calculated twice, once per chunk, as the sequence is 512 tokens long.
This highlights the trade-off between accuracy and efficiency: larger chunks improve results, but increase computation.

\paragraph{Learning at test time}
Finally, we evaluate Titans' ability to learn at test time. For this analysis, we split the CC-News dataset into 30\% training, 10\% validation, and 60\% testing. During this experiment, the parameters of the Transformer backbone are frozen, and only the neural memory weights are updated through the associative loss. After 50 epochs, performance remains stable, with variations in F1 and accuracy limited to the fourth decimal place; beyond this point, performance slowly deteriorates.
These findings suggest that memory updates alone are insufficient for meaningful test-time learning. One possible explanation for this is a mismatch between the frozen backbone input projections into key-value space and how the memory evolves.
Without joint adaptation, the integration of new information is limited.
Other factors, such as gating dynamics, memory capacity or interference effects, could also contribute.
This highlights the need for future work on coordinated adaptation between memory and backbone to realise the full potential of Titans for test-time learning.


\section{Conclusions}
In this work, we presented a reimplementation and critical analysis of Titans, a model combining attention mechanisms with a neural memory updated at test time.
Experiments on language modeling, recommendation, and time series forecasting confirm that the memory component mitigates information loss caused by chunking, enabling Titans to match or surpass competitive baselines.
However, the results also reveal some important trade-offs: larger chunks improve accuracy, but require substantial additional computing power. Conversely, memory updates alone proved insufficient for meaningful test-time learning when the backbone is frozen.
Overall, our work with Titans demonstrates both the potential and the challenges of test-time adaptation, revealing the need for more sophisticated integration strategies and a deeper investigation into the conditions under which neural memory can effectively enhance generalization beyond training.

\bibliography{Arxiv}

\end{document}